# Single Bitmap Block Truncation Coding of Color Images Using Hill Climbing Algorithm

Zhang Lige(张力戈)[*,**]   Qin Xiaolin(秦小林)[*,**]   Li Qing(李卿)[*,**]   Peng Haoyue(彭皓月)[*,**]   Hou Yu(侯屿)[*,**]

( [*]Chengdu Institute of Computer Applications, Chinese Academy of Sciences, Chengdu 610041, P. R. China )
( [**]School of Computer and Control Engineering, University of Chinese Academy of Sciences, Beijing 100049, P. R. China )

**Abstract**

Recently, the use of digital images in various fields is increasing rapidly. To increase the number of images stored and get faster transmission of them, it is necessary to reduce the size of these images. Single bitmap block truncation coding (SBBTC) schemes are compression techniques, which are used to generate a common bitmap to quantize the R, G and B planes in color image. As one of the traditional SBBTC schemes，weighted plane (W-plane) method is famous for its simplicity and low time consumption. However, the W-plane method also has poor performance in visual quality. This paper proposes an improved SBBTC scheme based on W-plane method using parallel computing and hill climbing algorithm. Compared with various schemes, the simulation results of the proposed scheme are better than that of the reference schemes in visual quality and time consumption.

**Key words:** block truncation coding, common bitmap, parallel computing, hill climbing algorithm.

## 0 Introduction

With the rapid development of the Internet, the number of digital images used by the public has become more and more huge. As a result, a large amount of storage space is required for digital images and the transmission of these images through limited bandwidth channel is time consuming. Therefore, image compression techniques have become one of the necessary way to overcome these issues.

Image compression refers to the process of representing an image in a compact way by reducing various redundancies in the image and it is generally classified into two categories: lossless compression and lossy compression [1]. Lossless compression techniques provide excellent visual quality of reconstructed images since there is no distortion introduced in their processes, but the compression ratio of them is low. So far, many lossless compression techniques have been proposed, such as run length coding [2], Huffman coding [3], arithmetic coding [4], etc. These techniques generally used in some special fields such as medical images and military images. Lossy compression provides high compression ratio and good visual quality of restored images. There are quite a few lossy compression techniques, such as vector quantization [5], fractal image compression [6], wavelet transform coding [7], block truncation coding [8], etc. These techniques are usually used to compress general-purpose digital images. As one of the lossy compression techniques, block truncation coding is famous for its simplicity.

Block truncation coding (BTC) first was proposed by Mitchel and Delp [8] for grayscale image compression. Due to the advantages such as low computational and fault tolerance, BTC has been also used extensively in many fields, such as image retrieval [9], data hiding [10] and image authentication [11], etc. In BTC, the grayscale image is divided into non-overlapping blocks and each block is encoded using a representative set of quantization values. Many schemes were proposed for improving the bit rate and visual quality of BTC. For instance, Mitchell et al. [12] proposed the absolute moment BTC, L. Hui [13] proposed the adaptive BTC, Haung et al. [14] proposed the hybrid BTC. The three typical schemes both have better visual quality than BTC. Later, BTC has been extended to the compression of color images.

A color image is typically composed of three channels, i.e., R, G and B. Apparently, color images contain a lot of data which are triple the data in grayscale images. Conventionally, BTC is used to code the three channels separately. However, this scheme has a low compression ratio. Therefore, it is important to utilize the correlation among the three channels to reduce the bit rate. To achieve this goal, some schemes have been proposed. In 1992, Wu and Coll [15] proposed the single bitmap BTC (SBBTC) scheme, which preserved the spatial details of the color image by using a common bitmap for the three channels. In this scheme, the weighted plane (W-plane) method which has low complexity and good performance regarding the mean square error (MSE) was proposed. Kurita and Otsu [16] used an adaptive one-bit vector quantization algorithm to generate the common bitmap in 1993. Tai et al. [17] proposed a new SBBTC scheme in 1997. The scheme used the Hopfield neural network to generate the common bitmap. Later, Tai et al. [18] proposed another new scheme using the genetic algorithm and the absolute moment BTC (GA-AMBTC) in 1998. The simulation results showed that this scheme had a better performance than the scheme they proposed earlier. In 2002, Chang and Wu [19] employed the gradual search algorithm with BTC to generate the common bitmap (GSBTC). This scheme has low complexity and acceptable visual quality. In 2013, Cui et al. [20] described a new scheme that incorporates the cat swarm optimization algorithm for generating the common bitmap (CSO-BTC). Recently, Li et al. [21] employed the binary ant colony optimization with BTC to generate the common bitmap (BACO-BTC). GA-AMBTC and CSO-BTC both are combined with the intelligence optimization algorithms, and they can be used to find the global optimal common bitmap after multiple iterations of the intelligence optimization algorithms they employed. However, the time consumption for them to find the global optimal common bitmap is huge. GSBTC and BACO-BTC have low computation complexity and can be employed to generate effective common bitmaps which have acceptable visual quality.

This paper proposes an improved scheme based on W-plane method. According to the characteristics of BTC, we employ the hill climbing algorithm to generate a near-optimal common bitmap and use parallel computing to reduce the time consumption. The simulation results show that our scheme has lower time consumption and better visual quality than that of other related schemes.

The remainder of this paper is organized as follows. Section 1 reviews some related works. Section 2 describes the proposed scheme in detail. Simulation results are given in Section 3 and conclusions are presented in Section 4.

# 1 Preliminaries

## 1.1 Block truncation coding

Block truncation coding (BTC) was proposed for compressing grayscale images. The basic steps of BTC are as follows.

Step 1. Partition grayscale image into nonoverlapping $m \times n$ blocks and the mean intensity of each block can be computed by

$$\bar{x} = \frac{1}{m \times n} \sum_{i=1}^{m} \sum_{j=1}^{n} x_{ij} \qquad (1)$$

where $x_{ij}$ is the value of pixel in each block at the position $(i, j)$.

Step 2. The bitmap B and two quantization values $x_H$ and $x_L$ are generated. Each pixel block is classified individually into two levels. For each pixel $x_{ij}$ in one block, if $x_{ij} < \bar{x}$, it is assigned to the low level and the value at the position $(i, j)$ in the bitmap is set to 0; otherwise, the pixel is assigned to the high level and the value at the position $(i, j)$ in the bitmap is set to 1. The two quantization values can be computed by

$$x_H = \frac{1}{q} \sum_{x_{ij} \geq \bar{x}} x_{ij} \qquad (2)$$

$$x_L = \frac{1}{m \times n - q} \sum_{x_{ij} < \bar{x}} x_{ij} \qquad (3)$$

where q is the number of pixels with a value higher than $\bar{x}$. The bitmap is generated by

$$b_{ij} = \begin{cases} 1 & x_{ij} \geq \bar{x} \\ 0 & x_{ij} < \bar{x} \end{cases} \qquad (4)$$

Step 3. Each block of the grayscale image M is reconstructed as follows:

$$m_{ij} = \begin{cases} x_H & b_{ij} = 1 \\ x_L & b_{ij} = 0 \end{cases} \qquad (5)$$

An example of this section is shown in **Fig.1.** **Fig.1(I)** is a $4 \times 4$ pixel block of original image. The mean intensity $\bar{x}$ and the two quantization values $x_H$ and $x_L$ of this block can be computed as 62, 86 and 22. **Fig.1(II)** shows the bitmap which is generated according to the Step 2. Then, the reconstructed block is created by Eq. (5) as the **Fig.1(III)** shows.

| 100 | 99 | 95 | 96 |
|---|---|---|---|
| 85 | 75 | 60 | 56 |
| 87 | 86 | 66 | 71 |
| 6 | 3 | 5 | 2 |

I

| 1 | 1 | 1 | 1 |
|---|---|---|---|
| 1 | 1 | 0 | 0 |
| 1 | 1 | 1 | 1 |
| 0 | 0 | 0 | 0 |

II

| 86 | 86 | 86 | 86 |
|---|---|---|---|
| 86 | 86 | 22 | 22 |
| 86 | 86 | 86 | 86 |
| 22 | 22 | 22 | 22 |

III

**Fig.1** The procedure of BTC, **I** pixel block of original image, **II** bitmap of the block, **III** the reconstructed pixel block

### 1.2 W-plane method

SBBTC was proposed to compress color images. Unlike traditional BTC of color images, SBBTC is used to compress the color image as one bitmap. W-plane method is one of the traditional SBBTC scheme.

In the W-plane method, the source color image is also partitioned into nonoverlapping m × n blocks. Let $R_{ij}, G_{ij}$ and $B_{ij}$ denote the pixel value of the R, G and B channels in each block, $x_{ij} = (R_{ij}, G_{ij}, B_{ij})$ denote the pixel vector. The weighted plane of each block is constructed by

$$w_{ij} = \frac{R_{ij} + G_{ij} + B_{ij}}{3} \tag{6}$$

and the mean intensity of each weighted plane can be computed by:

$$\bar{w} = \frac{1}{m \times n} \sum_{i=1}^{m} \sum_{j=1}^{n} w_{ij} \tag{7}$$

Then, the common bitmap of each block can be constructed according to $w_{ij}$ and $\bar{w}$. Each bit of the common bitmap is determined by

$$b_{ij} = \begin{cases} 1 & w_{ij} \geq \bar{w} \\ 0 & w_{ij} < \bar{w} \end{cases} \tag{8}$$

According to the common bitmap, the quantization vectors $x_H$ and $x_L$ can be computed:

$$x_H = \frac{1}{q} \sum_{b_{ij}=1} x_{ij} \tag{9}$$

$$x_L = \frac{1}{m \times n - q} \sum_{b_{ij}=0} x_{ij} \tag{10}$$

where $q$ stand for the number of $b_{ij}$ which equals 1.

Finally, each block of the color image C can be reconstructed as follows:

$$c_{ij} = \begin{cases} x_H & b_{ij} = 1 \\ x_L & b_{ij} = 0 \end{cases} \tag{11}$$

### 1.3 Hill climbing algorithm

Hill climbing algorithm attempts to compute a local optimal solution of the target function $f(x)$. The process of hill climbing algorithm can be summarized as follows:

Step 1. Select an initial point $x_0$ and record the optimal solution $x_{best} = x_0$. Have $T = N(x_0)$ denote the neighbourhood points of $x_0$.

Step 2. Stop the algorithm and output $x_{best}$ if $T = \emptyset$ ($\emptyset$ denote the empty set) or other stopping criterions are reached. Otherwise, go to Step 3.

Step 3. Select one point $x_n$ from $T$ and calculate $f(x_n)$. Set $x_{best} = x_n$, $T = T - x_n$ and go to Step 2 if $f(x_n)$ is larger than $f(x_{best})$. Otherwise, keep the value of $x_{best}$, set $T = T - x_n$ and go to Step 2.

## 2 The Proposed Scheme

In this paper, we propose an improved method based on w-plane method using parallel computing and hill climbing Algorithm. The proposed scheme consists of five phases as the **Fig.2** shows.

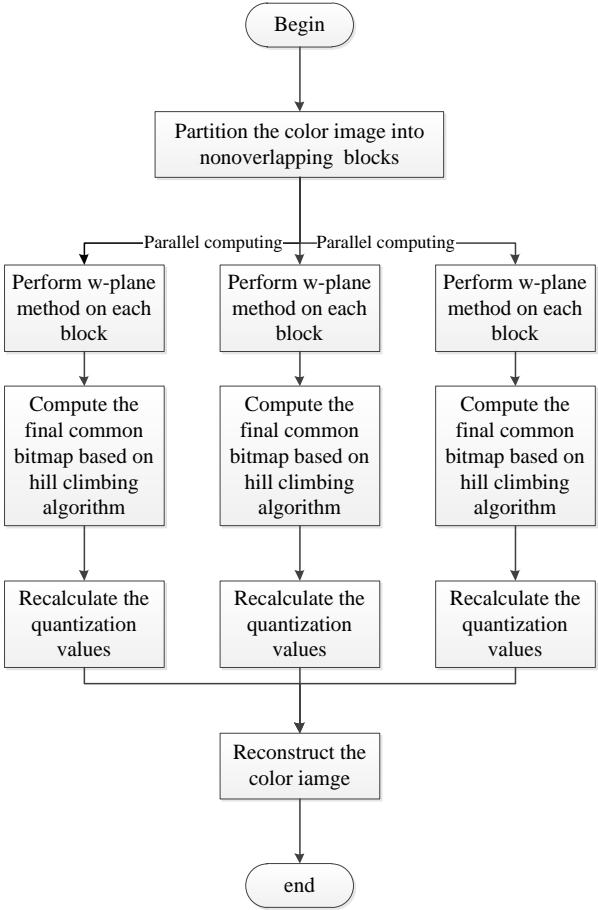

**Fig.2** Flowchart of the proposed scheme

## 2.1 Partition the color image into nonoverlapping blocks

In the proposed scheme, the original color image should be divided into a series of nonoverlapping blocks at first and the number of the nonoverlapping blocks is recorded as $B_{num}$. Since the subsequent operations for each block are identical and there is no interaction between these operations, these blocks can be processed independently. Therefore, the proposed scheme uses parallel computing for the operations of these blocks to fully utilize the CPU resources and reduce the time consumption.

Partition the blocks evenly into several parts according to the number of the cores on your machine so that we can use parallel computing. Assuming the machine has $C_{num}$ cores, then the number of blocks allocated to each core can be calculated:

$$CB_{num} = \frac{B_{num}}{C_{num}} \qquad (12)$$

To confirm the effectiveness of the parallel computing, we compare the time consumption of the parallel computing with that of no parallel computing for the proposed scheme. **Fig.3(I)** shows the result of comparison with $4 \times 4$ block size and **Fig.3(II)** shows the result of comparison with $8 \times 8$ block size. The blue line in **Fig.3** represents the proposed scheme using parallel computing and the red line stands for the proposed scheme without using parallel computing. It is obvious that the proposed scheme using the parallel computing has better time consumption.

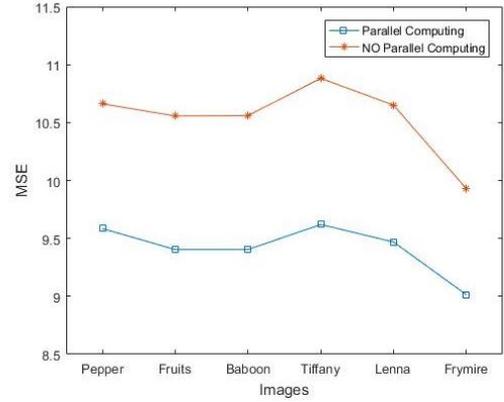

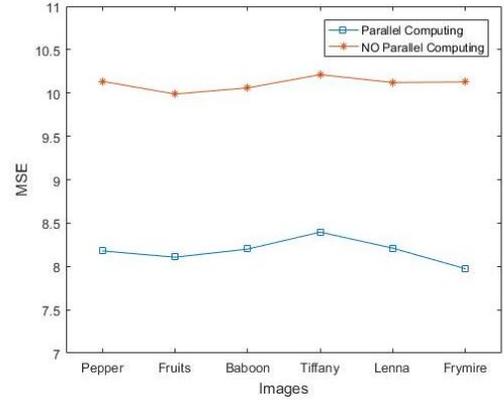

**Fig.3** Comparison for the time consumption of the parallel computing with that of no parallel computing

## 2.2 Perform W-plane method on each block

For each block, W-plane method is used to generate the initial common bitmap $B_{initial}$. The quantization values $x_{RH}, x_{RH}, x_{GH}, x_{GL}, x_{BH}, x_{BL}$ for the R, G and B channels are calculated according to $B_{initial}$, respectively, as is shown in Eq. (13).

$$x_{RH} = \frac{1}{q} \sum_{b_{ij}=1} R_{ij}$$

$$x_{RL} = \frac{1}{m \times n - q} \sum_{b_{ij}=0} R_{ij}$$

$$x_{GH} = \frac{1}{q} \sum_{b_{ij}=1} G_{ij}$$

$$x_{GL} = \frac{1}{m \times n - q} \sum_{b_{ij}=0} G_{ij} \quad (13)$$

$$x_{BH} = \frac{1}{q} \sum_{b_{ij}=1} B_{ij}$$

$$x_{BL} = \frac{1}{m \times n - q} \sum_{b_{ij}=0} B_{ij}$$

where $R_{ij}, G_{ij}$ and $B_{ij}$ are the pixel values at position $(i,j)$ of the R, G and B channels, $b_{ij}$ is the value at position $(i,j)$ of the initial common bitmap, $q$ refers to the number of $b_{ij}$ that equals 1, m and n stand for the size of each block.

## 2.3 Compute the final common bitmap based on hill climbing algorithm

The initial bitmap which is generated by w-plane method always results in poor visual quality. To overcome this issue, the proposed scheme uses hill climbing to generate the final bitmap.

Considering the huge time consumption when we optimize the entire bitmap at once using intelligent optimization algorithms and the independent effect when we change different bit of bitmap independently, the proposed scheme optimizes each bit of the bitmap separately. The detailed steps are as follows.

Step 1. Create an empty matrix BF. Calculate the initial MSE M using the initial bitmap $B_{initial}$ by Eq. (14).

$$MSE = \sum_{b_{ij}=0} (x_{ij} - x_L)^2 + \sum_{b_{ij}=1} (x_{ij} - x_H)^2 \quad (14)$$

where $x_{ij} = (R_{ij}, G_{ij}, B_{ij})$ stand for the pixel vector in the block, $x_H = (x_{RH}, x_{GH}, x_{BH})$ and $x_L = (x_{RL}, x_{GL}, x_{BL})$ are two vector consist of the six quantization values.

Step 2. Select one element $b_{mn}$ at position $(m,n)$ of $B_{initial}$ and initialize the optimal solution $b_{best}$ and target function $f(b_{mn})$ with $b_{mn}$ and M respectively.

Step 3. Find the neighbourhood points of $b_{mn}$.

Since the range of each element in $B_{initial}$ is {0,1}, there is only one neighbourhood point of $b_{mn}$. Hence, the neighbourhood point $t$ of $b_{mn}$ is calculated as follows:

$$t = 1 - b_{mn} \quad (15)$$

Step 4. Construct a temporary bitmap $BT$ by Eq. (16) and calculate $f(t)$ by Eq. (17). Note that there is only one difference between $BT$ and $B_{initial}$, which is located at position $(m,n)$.

$$bt_{ij} = \begin{cases} b_{ij} & (i,j) \neq (m,n) \\ t & (i,j) = (m,n) \end{cases} \quad (16)$$

$$f(t) = \sum_{bt_{ij}=1} (x_{ij} - x_H)^2 + \sum_{bt_{ij}=0} (x_{ij} - x_L)^2 \quad (17)$$

Step 5. Update $b_{best}$ as follows:

$$b_{best} = \begin{cases} b_{mn} & f(b_{mn}) \geq f(t) \\ T & f(b_{mn}) < f(t) \end{cases} \quad (18)$$

Step 6. Update the final bitmap $BF$ according to the position of $b_{mn}$ as follows:

$$bf_{mn} = b_{best} \quad (19)$$

Step 7. Repeat Step 2 to Step 6 until all the elements of $B_{initial}$ have been selected.

When all the elements of $B_{initial}$ have been selected, all the element of $BF$ are assigned. Up to this point, the final bitmap is obtained.

The main procedure of this part can be summarized as **Algorithm 1**.

---

**Algorithm 1.  Computation for the final common bitmap based on hill climbing algorithm**

**Input**: Block $C_{block}$ of the original color image, six quantization values $x_{RH}, x_{RH}, x_{GH}, x_{GL}, x_{BH}, x_{BL}$, the initial common bitmap $B_{initial}$ of $C_{block}$.
**Output**: Final common bitmap $BF$ of $C_{block}$.
1. Set **M** as the initial MSE;
2. Set **BF** as an empty matrix;
3. Set **BT** as a temporary bitmap
4. **for** i:=1 to m **do**
5.   **for** j:= 1 to n **do**
6.     $b_{best} \leftarrow b_{ij}$;
7.     $f(b_{ij}) \leftarrow M$;
8.     $t \leftarrow 1 - b_{ij}$;
9.     $bt_{ij} \leftarrow t$;
10.    $f(t) \leftarrow$ *new MSE computed according to BT*
11.    **if** $f(b_{ij}) \geq f(t)$ **then**
12.      $bf_{ij} \leftarrow b_{best}$
13.    **else**
14.      $b_{best} \leftarrow t$
15.      $bf_{ij} \leftarrow b_{best}$
16. **end do**

17.**end do**

There is an example in **Fig.4**. The selected element of initial common bitmap is $b_{42}$ and the target function $f(b_{42})$ is 1.2. First, find the neighbourhood point $t$ of $b_{42}$ and construct the temporary bitmap. Second, the result of target function $f(t)$ is 1.3 according to the temporary bitmap. Then, the optimal solution $b_{best}$ is updated to $b_{42}$ since $f(b_{42})$ is greater than $f(t)$,. In the end, the final common bitmap is updated by setting $bf_{42}$ to $b_{best}$.

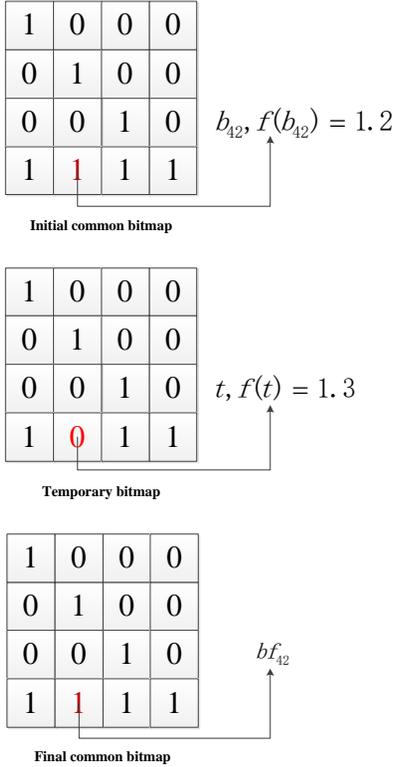

Fig.4 An example for Algorithm 1

### 2.4 Recalculate the quantization values

The quantization values are one of the main components for reconstructing the image in BTC. To reconstruct the color image, the quantization values need to be recalculated. According to the final common bitmap $BF$ which is generated in **Section 2.3**, the quantization values can be recalculated by Eq. (20).

$$\begin{aligned}
c_{RH} &= \frac{1}{q_c} \sum_{bf_{ij}=1} R_{ij} \\
c_{RL} &= \frac{1}{m \times n - q_c} \sum_{bf_{ij}=0} R_{ij} \\
c_{GH} &= \frac{1}{q_c} \sum_{bf_{ij}=1} G_{ij} \\
c_{GL} &= \frac{1}{m \times n - q_c} \sum_{bf_{ij}=0} G_{ij} \\
c_{BH} &= \frac{1}{q_c} \sum_{bf_{ij}=1} B_{ij} \\
c_{BL} &= \frac{1}{m \times n - q_c} \sum_{bf_{ij}=0} B_{ij}
\end{aligned} \quad (20)$$

where $c_{RH}, c_{RL}, c_{GH}, c_{GL}, c_{BH}, c_{BL}$ denote the six quantization values, $bf_{ij}$ stand for the value at position $(i,j)$ of the final common bitmap BF, $q_c$ is the number of $bf_{ij}$ which equals 1.

### 2.5 Reconstruct the color image

After getting the final common bitmap and the quantization values, the color image can be reconstructed as follows:

$$\begin{aligned}
R_{ij}^r &= \begin{cases} c_{RH} & bf_{ij}=1 \\ c_{RL} & bf_{ij}=0 \end{cases} \\
G_{ij}^r &= \begin{cases} c_{GH} & bf_{ij}=1 \\ c_{GL} & bf_{ij}=0 \end{cases} \\
B_{ij}^r &= \begin{cases} c_{BH} & bf_{ij}=1 \\ c_{BL} & bf_{ij}=0 \end{cases}
\end{aligned} \quad (21)$$

An example for this section is shown in **Fig.5**. Assume that the quantization values $c_{RH}, c_{RL}, c_{GH}, c_{GL}, c_{BH}, c_{BL}$ are equal to 235,226,182,156,255,250. According to the final common bitmap in **Fig.5**, the three channels can be reconstructed. R channel is reconstructed using $c_{RH}, c_{RL}$. G channel is reconstructed using $c_{GH}, c_{GL}$. B channel is reconstructed using $c_{BH}, c_{BL}$.

| Final common bitmap | | | | | R channel | | | |
|---|---|---|---|---|---|---|---|---|
| 1 | 0 | 0 | 0 | | 235 | 226 | 226 | 226 |
| 0 | 1 | 0 | 0 | | 226 | 235 | 226 | 226 |
| 0 | 0 | 1 | 0 | | 226 | 226 | 235 | 226 |
| 1 | 1 | 1 | 1 | | 235 | 235 | 235 | 235 |

G channel:

| 182 | 156 | 156 | 156 |
|---|---|---|---|
| 156 | 182 | 156 | 156 |
| 156 | 156 | 182 | 156 |
| 182 | 182 | 182 | 182 |

B channel:

| 255 | 250 | 250 | 250 |
|---|---|---|---|
| 250 | 255 | 250 | 250 |
| 250 | 250 | 255 | 250 |
| 255 | 255 | 255 | 255 |

**Fig.5** An example for the reconstruction of the color image

## 3 Simulation Results

The proposed scheme was simulated by MATLAB 2016a and the simulation environment was a PC with 8 GB RAM and two 3.4 GHz CPU. The color images used to evaluate the performance of the proposed scheme are show in **Fig.6**.

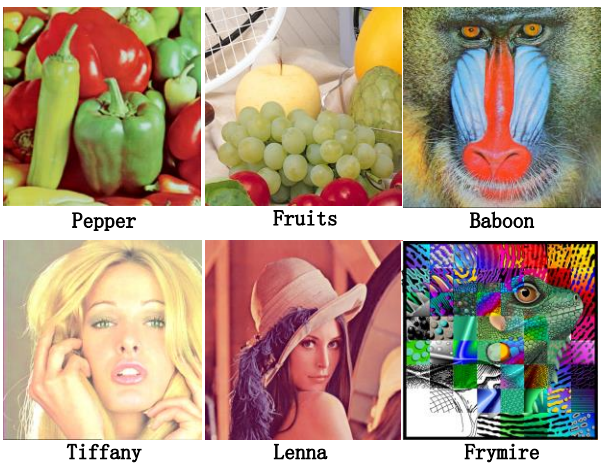

Pepper　　　Fruits　　　Baboon

Tiffany　　　Lenna　　　Frymire

**Fig.6**. Color test images

In **Table 1** and **Table 2**, we compare the proposed scheme with W-plane method. MSE is used to measure the performance of the two schemes, and it is defined as follows:

$$MSE = \frac{1}{m \times n} \sum_{i=1}^{m} \sum_{j=1}^{n} \frac{(R_{ij}^r - R_{ij})^2 + (G_{ij}^r - G_{ij})^2 + (B_{ij}^r - B_{ij})^2}{3}$$
(22)

where $R_{ij}, G_{ij}, B_{ij}$ stand for the pixel values of the three channels of the original color image at position $(i,j)$ and $R_{ij}^r, G_{ij}^r, RB_{ij}^r$ refer to the pixel values of the three channels of the reconstructed color image at position $(i,j)$, $m$ and $n$ are the size of the color image.

**Table 1** shows the comparison of MSE between the W-plane method [15] and the proposed method. The block size of this comparison is $4 \times 4$. It is seen that the proposed scheme has better MSE performance than W-plane method on each color image. **Table 2** shows the comparison of MSE with the block size of $8 \times 8$. It is obvious that the MSE performance of the proposed scheme is also better than W-plane method on each color image. These results illustrate that the improvement in the proposed scheme based on the W-plane method is efficient.

To further verify the validity of the proposed scheme, we compared the proposed scheme with other three related schemes, GA-AMBTC [18], GSBTC [19], BACO-BTC [21]. The parameters of GA-AMBTC are set according to Chang et al.'s works [18].

**Table 3** and **Table 4** show the comparison of MSE among the four different schemes. The block size of the color images used in **Table 3** and **Table 4** is $4 \times 4$ and $8 \times 8$, respectively. From the two tables, it is apparent that the MSE of the proposed scheme is lower than that of the other three related schemes, which means that our scheme is better.

**Table 5** and **Table 6** show the comparison of structural similarity(SSIM) index among the four different schemes with block sizes of $4 \times 4$ and $8 \times 8$, respectively. SSIM is a method for measuring the similarity between two images. It can be calculated by Eq. (23).

$$SSIM(X,Y) = \frac{(2\mu_X \mu_Y + C_1)(2\sigma_{XY} + C_2)}{(\mu_X^2 + \mu_Y^2 + C_1)(\sigma_X^2 + \sigma_Y^2 + C_2)}$$
(23)

where X is the original color image, Y is the reconstructed color image, $\mu_X$ and $\mu_Y$ refer to the average of the two images, $\sigma_{XY}$ is the covariance of the two images, $\sigma_X$ and $\sigma_Y$ stand for the variance of X and Y respectively, $C_1$ and $C_2$ are two default constants.

In the two tables, the proposed scheme has higher SSIM values than the other three schemes. This result confirms that the proposed scheme has better visual quality.

Table 1. MSE for images reconstructed by W-plane method and the proposed scheme with block size 4 × 4

| Images | W-plane-BTC [15] | Proposed scheme |
|---|---|---|
| Pepper | 11.1420 | 10.4734 |
| Fruits | 19.5508 | 16.8978 |
| Baboon | 95.9592 | 90.7861 |
| Tiffany | 19.2256 | 12.7837 |
| Lenna | 19.9605 | 18.7425 |
| Frymire | 254.1143 | 213.5586 |

Table 2. MSE for images reconstructed by W-plane method and the proposed scheme with block size 8 × 8

| Images | W-plane-BTC [15] | Proposed scheme |
|---|---|---|
| Pepper | 32.8668 | 29.8284 |
| Fruits | 35.9711 | 30.6846 |
| Baboon | 145.6842 | 135.5584 |
| Tiffany | 31.7653 | 22.9964 |
| Lenna | 38.6159 | 35.6684 |
| Frymire | 468.8183 | 394.8452 |

Table 3. MSE for images reconstructed by the four schemes with block size 4 × 4

| Image | GA-AMBTC [18] | GSBTC [19] | BACO-BTC [21] | Proposed scheme |
|---|---|---|---|---|
| Pepper | 10.6488 | 13.2820 | 10.5253 | 10.4734 |
| Fruits | 18.4856 | 19.1409 | 16.9901 | 16.8978 |
| Baboon | 95.1157 | 98.5005 | 90.7963 | 90.7861 |
| Tiffany | 13.5013 | 24.1907 | 14.9005 | 12.7837 |
| Lenna | 19.6789 | 20.1871 | 18.7826 | 18.7425 |
| Frymire | 239.3385 | 336.0331 | 226.0877 | 213.5586 |

Table 4. MSE for images reconstructed by the four schemes with block size 8 × 8

| Image | GA-AMBTC [18] | GSBTC [19] | BACO-BTC [21] | Proposed scheme |
|---|---|---|---|---|
| Pepper | 32.0923 | 35.4107 | 29.9243 | 29.8284 |
| Fruits | 34.4711 | 35.3892 | 31.2399 | 30.6846 |
| Baboon | 145.9833 | 146.2950 | 136.1672 | 135.5584 |
| Tiffany | 33.2753 | 35.0836 | 26.0522 | 22.9964 |
| Lenna | 38.4107 | 36.5837 | 35.4927 | 35.6684 |
| Frymire | 447.5942 | 551.3695 | 414.1567 | 394.8452 |

Table 5. SSIM for images reconstructed by the four schemes with block size 4 × 4

| Image | GA-AMBTC [18] | GSBTC [19] | BACO-BTC [21] | Proposed scheme |
|---|---|---|---|---|
| Pepper | 0.9924 | 0.9911 | 0.9925 | 0.9926 |
| Fruits | 0.9793 | 0.9790 | 0.9804 | 0.9806 |
| Baboon | 0.9007 | 0.9029 | 0.9059 | 0.9060 |
| Tiffany | 0.9854 | 0.9758 | 0.9838 | 0.9865 |
| Lenna | 0.9859 | 0.9855 | 0.9867 | 0.9867 |
| Frymire | 0.9409 | 0.9218 | 0.9449 | 0.9471 |

Table 6. SSIM for images reconstructed by the four schemes with block size 8 × 8

| Image | GA-AMBTC [18] | GSBTC [19] | BACO-BTC [21] | Proposed scheme |
|---|---|---|---|---|
| Pepper | 0.9777 | 0.9768 | 0.9790 | 0.9791 |
| Fruits | 0.9607 | 0.9606 | 0.9636 | 0.9641 |
| Baboon | 0.8539 | 0.8598 | 0.8640 | 0.8645 |
| Tiffany | 0.9694 | 0.9639 | 0.9740 | 0.9769 |
| Lenna | 0.9748 | 0.9760 | 0.9767 | 0.9767 |
| Frymire | 0.8837 | 0.8633 | 0.8927 | 0.8970 |

**Table 7** and **Table 8** show the comparison of time consumption among the four different schemes. To reflect the differences in time consumption among the four methods more intuitively, the size of the test images is set to 1024 × 1024. The block size in **Table 7** and **Table 8** is 4 × 4 and 8 × 8 respectively.

Consistent with the analysis in the introduction, the time consumption of GA-AMBTC in the two tables is significantly higher, and the time consumption of GSBTC and BACO-BTC is relatively small. It is obvious that the time consumption of the proposed scheme is lower than the other three related schemes. Therefore, the proposed scheme is more efficient.

Table 7. Time consumption (second) for images reconstructed by the four schemes with block size $4 \times 4$

| Image | GA-AMBTC [18] | GSBTC [19] | BACO-BTC [21] | Proposed scheme |
|---|---|---|---|---|
| Pepper | 666.9923 | 14.3152 | 9.8987 | 9.5844 |
| Fruits | 665.5733 | 13.7884 | 9.6280 | 9.4037 |
| Baboon | 678.5988 | 14.3051 | 9.4130 | 9.4043 |
| Tiffany | 667.8971 | 15.5259 | 9.7571 | 9.6200 |
| Lenna | 671.9080 | 14.2471 | 9.6702 | 9.4689 |
| Frymire | 677.2049 | 12.6800 | 9.6017 | 9.0140 |

Table 8. Time consumption (seconds) for images reconstructed by the four schemes with block size $8 \times 8$

| Image | GA-AMBTC [18] | GSBTC [19] | BACO-BTC [21] | Proposed scheme |
|---|---|---|---|---|
| Pepper | 331.2923 | 12.4233 | 10.2679 | 8.1790 |
| Fruits | 329.5832 | 11.4148 | 10.2335 | 8.1065 |
| Baboon | 330.6240 | 11.7746 | 10.2838 | 8.2000 |
| Tiffany | 328.8598 | 13.5416 | 10.3134 | 8.3956 |
| Lenna | 330.8747 | 12.2374 | 10.4508 | 8.2104 |
| Frymire | 329.8919 | 10.2554 | 10.2649 | 7.9732 |

## 4 Conclusion

In this paper, an efficient color image scheme based on W-plane method is proposed. According to the characteristics of BTC, we use parallel computing to reduce the time consumption. To obtain a near-optimal common bitmap, hill climbing algorithm is employed in this scheme. The simulation results show that the visual quality of the proposed scheme is better and the time consumption of the proposed scheme is lower than the related schemes. These results indicate that our scheme is efficient for color image compression and beneficial to store and transmit color images. In the future, this scheme can be further improved and used in other fields.

**References**
[1] Rabbani M, Jones P W. Digital image compression techniques[M]. SPIE Press, 1991.
[2] Pountain D. Run-length encoding[J]. Byte, 1987, 12(6): 317-319.
[3] Huffman D A. A method for the construction of minimum-redundancy codes[J]. Proceedings of the IRE, 1952, 40(9): 1098-1101.
[4] Witten I H, Neal R M, Cleary J G. Arithmetic coding for data compression[J]. Communications of the ACM, 1987, 30(6): 520-540.
[5] Goldberg M, Boucher P, Shlien S. Image compression using adaptive vector quantization[J]. IEEE Transactions on Communications, 1986, 34(2): 180-187.
[6] Fisher Y. Fractal image compression[J]. Fractals, 1994, 2(03): 347-361.
[7] DeVore R A, Jawerth B, Lucier B J. Image compression through wavelet transform coding[J]. IEEE Transactions on information theory, 1992, 38(2): 719-746.
[8] Delp E, Mitchell O. Image compression using block truncation coding[J]. IEEE transactions on Communications, 1979, 27(9): 1335-1342.
[9] Guo J M, Prasetyo H. Content-based image retrieval using features extracted from halftoning-based block truncation coding[J]. IEEE Transactions on image processing, 2015, 24(3): 1010-1024.
[10] Sun W, Lu Z M, Wen Y C, et al. High performance reversible data hiding for block truncation coding compressed images[J]. Signal, Image and Video Processing, 2013, 7(2): 297-306.
[11] Lin C C, Huang Y, Tai W L. A novel hybrid image authentication scheme based on absolute moment block truncation coding[J]. Multimedia tools and applications, 2017, 76(1): 463-488.
[12] Lema M, Mitchell O. Absolute moment block truncation coding and its application to color images[J]. IEEE Transactions on communications, 1984, 32(10): 1148-1157.
[13] Hui L. An adaptive block truncation coding algorithm for image compression[C]//Acoustics, Speech, and Signal Processing, 1990. ICASSP-90., 1990 International Conference on. IEEE,

1990: 2233-2236.
[14] Haung C S, Lin Y. Hybrid block truncation coding[J]. IEEE Signal Processing Letters, 1997, 4(12): 328-330.
[15] Wu Y, Coll D C. Single bit-map block truncation coding of color images[J]. IEEE Journal on Selected Areas in Communications, 1992, 10(5): 952-959.
[16] Kurita T, Otsu N. A method of block truncation coding for color image compression[J]. IEEE transactions on communications, 1993, 41(9): 1270-1274.
[17] Tai S C, Lin Y C, Lin J F. Single bit-map block truncation coding of color images using a Hopfield neural network[J]. Information Sciences, 1997, 103(1-4): 211-228.
[18] Tai S C, Chen W J, Cheng P J. Genetic algorithm for single bit map absolute moment block truncation coding of color images[J]. Optical Engineering, 1998, 37(9): 2483-2491.
[19] Chang C C, Wu M N. An Algorithm for Color Image Compression Base on Common Bit Map Block Truncation Coding[C]//JCIS. 2002: 964-967.
[20] Cui S Y, Wang Z H, Tsai P W, et al. Single bitmap block truncation coding of color images using cat swarm optimization[M]//Recent advances in information hiding and applications. Springer, Berlin, Heidelberg, 2013: 119-138.
[21] Li Z, Jin Q, Chang C C, et al. A Common Bitmap Block Truncation Coding for Color Images Based on Binary Ant Colony Optimization[J]. KSII Transactions on Internet and Information Systems (TIIS), 2016, 10(5): 2326-2345.